\documentclass[letterpaper, 10 pt, journal, twoside]{ieeetran}

\IEEEoverridecommandlockouts                              

\usepackage{graphicx} 
\usepackage{amsmath} 
\usepackage{xcolor}
\usepackage{ amssymb }
\usepackage{booktabs}
\usepackage{siunitx}
\usepackage{hyperref}

\DeclareMathOperator*{\argmin}{arg\,min}
\newcommand{\Loss}{\mathcal{L}}

\title{
A Kinematic Bottleneck Approach For Pose Regression of Flexible Surgical Instruments directly from Images}

\author{Luca Sestini$^{1,2}$, Benoit Rosa$^{1}$, Elena De Momi$^{2}$, Giancarlo Ferrigno$^{2}$ and Nicolas Padoy$^{1,3}$ 
\thanks{Manuscript received: October 15, 2020; Revised: January 10, 2021; Accepted: February 8, 2021.}
\thanks{This paper was recommended for publication by Editor Eric Marchand upon evaluation of the Associate Editor and Reviewers' comments.\\
This work was supported by the ATLAS project. This project has received funding from the European Union’s Horizon 2020 research and innovation programme under the Marie Sklodowska-Curie grant agreement No. 813782.\\This work is also partially supported by French State Funds managed by the Agence Nationale de la Recherche (ANR) through the Investissements d’Avenir Program under Grant ANR-11-LABX-0004 (Labex CAMI) and Grant ANR-10-IAHU-02 (IHU-Strasbourg)} 
\thanks{$^{1}$ Luca Sestini, Benoit Rosa and Nicolas Padoy are with ICube, University of Strasbourg, CNRS, IHU Strasbourg, France {\tt\footnotesize \{sestini,b.rosa,npadoy\}@unistra.fr}}%
\thanks{$^{2}$ Luca Sestini, Elena De Momi and Giancarlo Ferrigno are with Department of Electronics, Information and Bioengineering, Politecnico di Milano, Milano, Italy {\tt\footnotesize \{elena.demomi,giancarlo.ferrigno\}@polimi.it}}
\thanks{$^{3}$ Nicolas Padoy is with IHU Strasbourg, Strasbourg, France.}
\thanks{Digital Object Identifier (DOI): see top of this page.}
}

\begin{document}

\maketitle

\markboth{IEEE Robotics and Automation Letters. Preprint Version. Accepted February, 2021}
{Sestini \MakeLowercase{\textit{et al.}}: Kinematic Bottleneck Approach for Surgical Tools Pose Regression} 

\begin{abstract}

3-D pose estimation of instruments is a crucial step towards automatic scene understanding in robotic minimally invasive surgery. Although robotic systems can potentially directly provide joint values, this information is not commonly exploited inside the operating room, due to its possible unreliability, limited access and the time-consuming calibration required, especially for continuum robots. For this reason, standard approaches for 3-D pose estimation involve the use of external tracking systems. Recently, image-based methods have emerged as promising, non-invasive alternatives. While many image-based approaches in the literature have shown accurate results, they generally require either a complex iterative optimization for each processed image, making them unsuitable for real-time applications, or a large number of manually-annotated images for efficient learning. In this paper we propose a self-supervised image-based method, exploiting, at training time only, the imprecise kinematic information provided by the robot.
In order to avoid introducing time-consuming manual annotations, the problem is formulated as an auto-encoder, smartly \textit{bottlenecked} by the presence of a physical model of the robotic instruments and surgical camera, forcing a separation between image background and kinematic content. Validation of the method was performed on \textit{semi-synthetic}, \textit{phantom} and \textit{in-vivo} datasets, obtained using a flexible robotized endoscope, showing promising results for real-time image-based 3-D pose estimation of surgical instruments.

\end{abstract}


  \begin{figure}[t]
    \centering
      \includegraphics[width=3in]{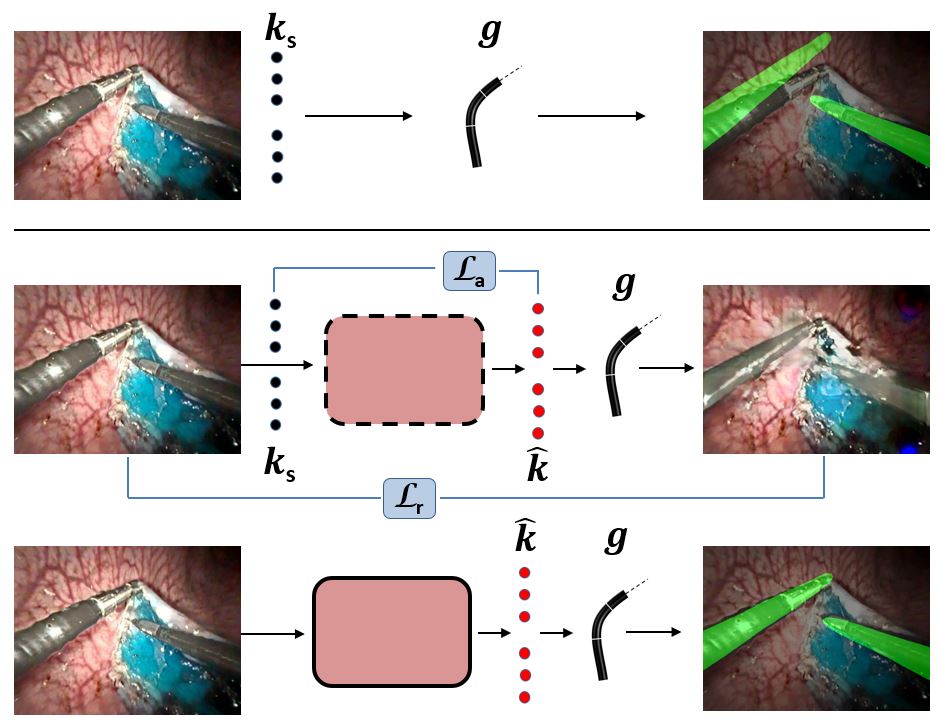}
      \caption{Illustration of the proposed approach. Top row: example of an image with associated kinematics $\boldsymbol{k_s}$, measured by the robotic system. Using a  model of the robotic instruments, $\boldsymbol{k_s}$ can be mapped to their 3-D shape $g$, and then reprojected on the image plane. The projected silhouette shows the inaccuracy of $\boldsymbol{k_s}$, due to the strong tool-tissue interaction; middle row: our model (red box) is trained by means of a self-supervised reconstruction loss ($\Loss_r$) and a weakly-supervised loss ($\Loss_a$) provided by the uncertain kinematics $\boldsymbol{k_s}$, without any manual supervision; bottom row: at inference time the model predicts kinematics from image-information only, with a superior accuracy with respect to the measured kinematics $\boldsymbol{k_s}$.}
      \label{init}
\end{figure}

 \begin{figure*}     
  \includegraphics[width=\textwidth]{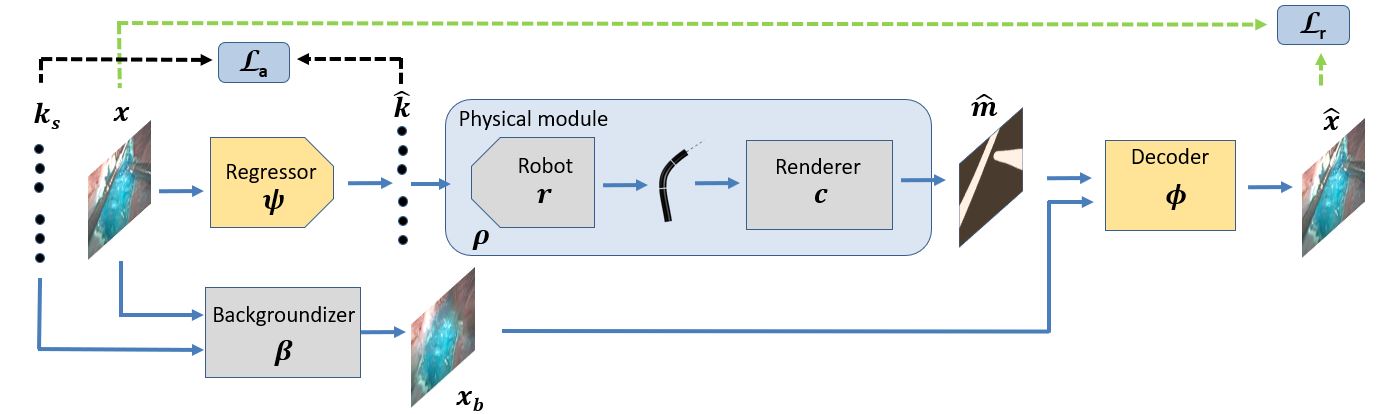}
  \caption{Full training architecture: given an input image $x$ and the corresponding imprecise measured kinematics $\boldsymbol{k_s}$, the training architecture forces a separation between the appearance of the image and its kinematic content, while keeping a self-supervised formulation. In the bottom branch, the module $\beta$ transforms $x$ into a \textit{backgroundized} version of itself $x_b$. In the top branch, the regressor $\psi$ reduces $x$ to the estimated kinematic configuration $\boldsymbol{\hat{k}}$, which is then converted by the physical model $\rho$ of the instruments and surgical camera in the binary silohuette projection $\hat{m}$ of the instruments on the image plane. A decoder $\phi$ tries to reconstruct the input image $x$ from $x_b$ and $\hat{m}$. The model is trained by means of the reconstruction loss $\Loss_r$, helped by the auxiliary loss $\Loss_a$. The \textit{backgroundizer} and the \textit{physical module} are trained in advance and frozen during the training of regressor and decoder.}
  \label{arch}
\end{figure*}

\section{INTRODUCTION}

\label{section:intro}
\IEEEPARstart{R}{obotic} Minimally Invasive Surgery (RMIS) is becoming a standard paradigm across different surgical specialties \cite{palep2009robotic}, providing surgeons with a significantly less invasive method of accessing distal sites, together with a level of precision and dexterity comparable to open surgery. Robotic systems are generally integrated with computer assistance, providing the surgeons with rich information about the intervention. A necessary step towards complete integration of computer assistance and RMIS is represented by the automatic localization of tools in the surgical scene, both at image level and in the 3-D space \cite{maier2017surgical}. Tool localization at image level offers a wide range of applications, from workflow analysis to tool subtraction in panorama reconstruction, augmented reality and visual servoing \cite{munzer2018content}. At 3-D level, localization of the instruments can be exploited to provide dynamic motion constraints, especially in the context of flexible robots \cite{azimian2010constrained}\cite{omisore2020motion}, in haptic-feedback applications \cite{okamura2004methods}, in tumor localization via remote palpation \cite{talasaz2012remote}, to analyze surgeons skills and in general for the above-mentioned 2-D applications, via reprojection onto the image space. In this paper we concentrate on the 3-D pose estimation problem, focusing, but not limiting to, continuum robots.\\ 
\indent Continuum robots have been developed to access the most challenging surgical sites in a minimally invasive way \cite{burgner2015continuum}. Compared to classical rigid articulated robots, continuum robots bend continuously when actuated, resulting in increased dexterity, at the price of more challenging control and modelling \cite{shi2016shape}. Recently developed robotic systems can potentially provide the kinematic information associated with each joint, measuring it at the motor side. However, this information is often unreliable due to two main factors: 1) tool-tissue interactions, which can modify the shape of a flexible instruments; 2) un-modelled non-linearities (e.g. cable friction, slack in instrument channel, backlash), which can lead to loss of motion between the motors and the instruments; these two factors can lead to a non-deterministic mapping between measured kinematics and actual robot configuration \cite{de2013introducing} (fig. \ref{init}, top row). Some of the mentioned phenomena depend on the type of instruments used, as well as their initial positioning, usually performed manually, making calibration a necessary, time-consuming step before every procedure.
For those reasons accessing robot kinematics in real-time inside the operating room (OR) is uncommon \cite{allan20183}, even when access to the robot API is granted. 
Standard methods for robots shape estimation involve attaching external sensors to the instruments (mainly electromagnetic, optical markers, Fiber Bragg Gratings), using trackers to estimate the shape \cite{shi2016shape}\cite{chmarra2007systems}\cite{gadwe2018real}\cite{wang2020image}. However, relying on external sensors involves several undesirable actions to be taken, such as modifying the instrument design to fit the sensors, introducing trackers in the OR and in general, altering the normal surgical workflow. For this reason, vision-based marker-less solutions have recently gained strong interest \cite{bouget2017vision}.\\
\indent The problem of estimating robot instruments pose from surgical camera images is generally posed as maximizing the agreement between a parametrized model of the instruments and features extracted from the image. Statistical region-based methods formulate the problem as estimating the pose parameters such that the corresponding silhouette regions, determined by the projection of a 3-D model of the instruments on the camera plane, match the silhouette regions estimated from a pixel-wise classification of the image. Works on rigid articulated robots, based on this approach, have shown good accuracy \cite{allan2015image}\cite{allan20183}. However, those methods require a per-frame optimization, which limits their real-time applicability. In addition, their accuracy is dependent on the quality of the segmentation model: the proposed Random Forest algorithms, initialized from a single frame, may not be able to handle a high variability of the data appearance (although online learning is implemented in \cite{allan20183}), and more powerful models may be needed, requiring, in turn, a large amount of annotated data for training. A different approach for surgical instruments pose estimation has been proposed in \cite{laina2017concurrent}\cite{kurmann2017simultaneous}\cite{du2018articulated}, which used deep learning models to directly localize the joint position on the image. Despite achieving high localization accuracy, those methods are limited to 2-D space, and require manual annotations for training. In general, for successful generalizability, supervised machine learning approaches require a great amount of labelled data acquired through manual annotation, which is both costly and time-consuming.\\
 \indent Several works in the computer-assisted interventions community have recently explored ways to mitigate the need for such amount of manual annotations, mainly focusing on two approaches: developing strategies to learn from synthetic data \cite{mu2020learning}\cite{heimann2014real}\cite{pakhomov2020towards} and reducing the amount of labels needed for training \cite{yu2018learning}\cite{ross2018exploiting}. In \cite{nwoye2019weakly}, Nwoye et al. explored weak-supervision, using cheap frame-level annotations on tool presence, to train a model for 2-D tool tracking. In \cite{tanwani2020motion2vec}, Tanwani et al. developed a semi-supervised approach based on surgical actions grouping, in order to obtain meaningful embedded representations of endoscopic video segments, showing promising results for downstream tasks as segmentation and pose imitation. A side of weak-supervision that has been marginally explored, especially in the surgical domain, is represented by the use of easy-to-obtain, but possibly inaccurate information. This paper demonstrates the benefits of using inaccurate kinematics information for self-supervision during training.
 
 In this work we address the problem of estimating the 3-D pose of flexible surgical instruments and propose a new self-supervised, image-based approach, which:
\begin{itemize}
    \item{can estimate the \textit{full} 3-D pose of instruments, parametrized by means of their joint values}, from images only, with no additional optimization required at inference time, making it easily suitable for real-time applications;
    \item{is trained end-to-end in a self-supervised way, exploiting an auto-encoder formulation, smartly \textit{bottlenecked} by the presence of a physical model of instruments and surgical camera}, in order to extract their kinematics;
    \item{leverages, at training time only, the rough localization provided by the uncertain measured kinematics, without requiring any manual annotation.}
\end{itemize}

\section{METHOD}
\label{section:method}
Our goal is to train a regressor model to estimate the pose of surgical instruments from image data acquired by the surgical camera.  Differently from standard methods as \cite{allan20183}, where the estimated pose is represented by the position and orientation of the end-effector, the regressor is trained to directly estimate the value of the $n$ kinematic joints $\boldsymbol{k} = \{k_0, k_1, ..., k_{n-1}\}$ of the instruments. The joint values, combined with a geometric forward kinematic model of the instruments, allow to reconstruct their \textit{full} 3-D shape.\\
The problem is formulated as learning the set of parameters $\boldsymbol{\theta_{\psi}}$ of the regressor model $\psi:x \rightarrow \boldsymbol{\hat{k}}$ that maps an image $x$, containing the instruments in a configuration represented by the set of ground truth joint values $\boldsymbol{k} = \{k_0, k_1, ..., k_{n-1}\}$, to the set of estimated joint values $\boldsymbol{\hat{k}}$, such that:

\begin{equation}
    \boldsymbol{\hat{\theta}_{\psi}} = \argmin_{\boldsymbol{\theta_{\psi}}}{(\boldsymbol{\psi(\boldsymbol{\theta_{\psi}},x)}- \boldsymbol{k)}}.
\end{equation}

However, $\boldsymbol{k}$ is not known in practice. The measured joint values $\boldsymbol{k_s}$, recorded by the robotic system, are generally inaccurate, due to tool-tissue interactions and possible un-modelled non-linearities, thus they cannot be directly used as supervision signal. In order to avoid relying on time-consuming and expensive manual annotations, we propose a training architecture built as an auto-encoder, trained by reconstructing the input image $x$ as output of the decoder model $\phi$, defined by the set of parameters $\boldsymbol{\theta_{\phi}}$. The complete training architecture can be observed in fig. \ref{arch}. The architecture, inspired by \cite{jakab2020self}, is based on the idea of separating the appearance of the image from its kinematic content. In order to do that, the architecture presents two separate branches. The \textit{appearance} branch consists of a module $\beta$, called \textit{backgroundizer module}, which converts the input image $x$ to an approximate background-only version of itself $x_b$. The \textit{content} branch contains the kinematic regressor $\psi$. However, without further constraints, the kinematic extraction would not be possible: the regressed vector $\boldsymbol{\hat{k}}$ would not necessarily contain only kinematic information, and it would not necessarily be physically meaningful (i.e.: each regressed value corresponding to a different specific joint). For this reason, the regressor $\psi$ is followed by a physical model of robotic instruments and surgical camera. This module $\rho$, called \textit{physical module}, maps the estimated kinematics $\boldsymbol{\hat{k}}$ to the instruments 3-D shape, through a forward kinematic model of the instruments, and reprojects it to the camera plane as a binary mask $\hat{m}$, representing the instruments in the image space, according to the estimated kinematics $\boldsymbol{\hat{k}}$. The \textit{physical module} $\rho$ has the pivotal role of a \textit{kinematic bottleneck}: because of the way it processes the low dimensional vector $\boldsymbol{\hat{k}}$, it gives it the explicit meaning of ``kinematics'', forcing, in turn, $\psi$ to learn the expected regression transformation, while keeping the self-supervised formulation of the problem. The outputs of the two branches, $x_b$ and $\hat{m}$, are finally fed to the decoder $\phi$, which tries to reconstruct $x$ from the combination of the two. The \textit{backgroundizer} and the \textit{physical} modules are discussed in sections \ref{back_module} and \ref{phys_module}.

 \begin{figure}[tpb]
    \centering
      \includegraphics[width=3in]{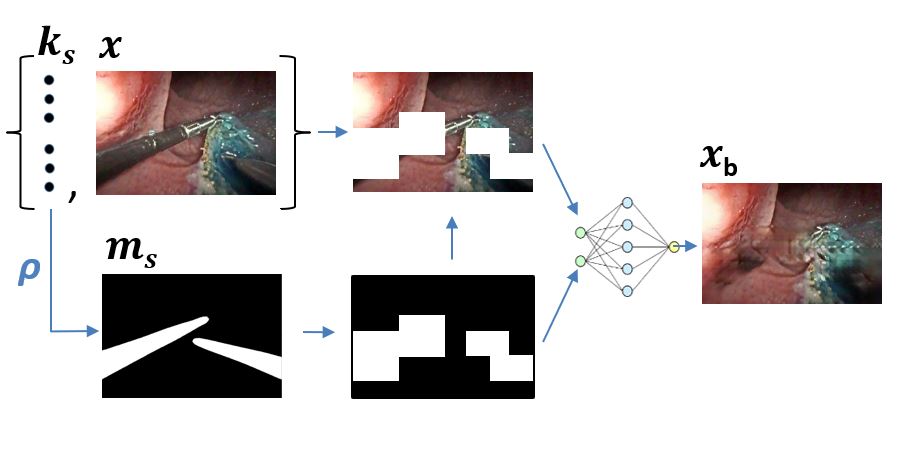}
      \caption{\textit{Backgroundizer module} $\beta$: the imprecise kinematics $\boldsymbol{k_s}$ is converted to the binary projection representation $m_s$, through a robot-renderer model equivalent to $\rho$. $m_s$ is then expanded to account for uncertainties and used to mask the image $x$ associated with $k_s$, which is then fed to an inpainting network which produces $x_b$, the \textit{backgroundized} version of $x$.}
      \label{back}
\end{figure}

\subsection{Backgroundizer module: an inpainting problem}
\label{back_module}
In order to make the reconstruction task feasible, the decoder must be provided with information about the appearance of the image. In \cite{jakab2020self}, in human pose-estimation context, Jakab et al. provide this information by feeding a second image $x'$ to the decoder, shifted in time with respect to $x$ (with $x$, $x'$ belonging to the same video). This approach is effective as long as $x'$ contains the same exact background as $x$ and the subject in it has a different pose than the one the model is trying to regress. Unfortunately, none of the hypotheses can be verified in a complex surgical scenario: the robotic instruments can remain in the same configuration for relatively long periods during a surgery, and the background appearance is constantly modified by the direct and indirect action of the instruments (pulling tissue, bleeding, smoke etc.) and by the movement of the camera. To address this issue, we introduce a novel \textit{backgroundizer module} $\beta$ that informs the decoder about the aspect of the image to reconstruct by producing the \textit{backgroundized} image $x_b$ as:
\begin{equation}
    x_b = \beta(x,\boldsymbol{k_s}).
\end{equation}
The image $x_b$ is obtained by exploiting the rough instruments localization provided by $\boldsymbol{k_s}$, accounting for its uncertainty. 
In order to leverage the imprecise information provided by the input kinematics $\boldsymbol{k_s}$, we first map it to the corresponding binary mask $m_s$ through a robot-renderer model,  equivalent to $\rho$. We take into account the possible error in  $\boldsymbol{k_s}$ by expanding the binary mask $m_s$, in order to cover a greater space in the image. To do that, we simply consider the tool presence in quadrants obtained by overlapping a grid with {$a\times b$} resolution on the mask $m_s$, and assigning to the whole quadrants a value of $1$ if at least one pixel of $m_s$ inside is $1$. The input image $x$ and the expanded mask are then processed to obtain the \textit{backgroundized} image $x_b$: the problem is formulated as image inpainting from block occlusion \cite{elharrouss2019image}. To solve it we implemented and trained a neural network model based on partial convolutions, following the implementation proposed in \cite{liu2018image}. The model is trained using a perceptual loss function, as defined in \cite{gatys2015neural}. The full pipeline for the \textit{backgroundizer module} can be observed in fig. \ref{back}.

\subsection{Physical module}
\label{phys_module}
The \textit{physical module} $\rho$, following the regressor $\psi$, has the crucial role of \textit{kinematic bottleneck}, giving $\boldsymbol{\hat{k}}$ the explicit meaning of ``kinematics'', and forcing $\psi$ to properly learn the kinematic regression task, while keeping the self-supervised formulation of the problem. This module consists of a geometrical, forward-kinematic model of the instruments, followed by a renderer (pinhole model of the surgical camera and rasterizer). The geometric forward kinematic model of the instruments is defined as the mapping $r: \boldsymbol{k} \rightarrow g$, with $g$ being the 3-D shape of the instruments, referred to the instruments reference frame $\{r_{rf}\}$. Knowing the parameters of the rigid transformation $T$ between $\{r_{rf}\}$ and the robot-mounted camera reference frame $\{c_{rf}\}$, and its calibration matrix, the instruments 3-D shape $g$ can be projected on the camera plane and rasterized in pixel coordinates. The full rendering operation is defined as $c: g \rightarrow \hat{m}$, with $\hat{m}$ being the binary instruments segmentation mask, obtained binarizing the projection of the 3-D shape of the instruments $g$ on the camera plane.

The full problem can be so formulated as:
\begin{equation}
    \boldsymbol{\hat{\theta}_{\psi}},\boldsymbol{\hat{\theta}_{\phi}} = \argmin_{\boldsymbol{\theta_{\psi}},\boldsymbol{\theta_{\phi}}}{\phi(\boldsymbol{\theta_{\phi}},\rho(\psi(\boldsymbol{\theta_{\psi}},x)),\beta(x,\boldsymbol{k_s})) - x}.
\label{eq:autoencod}
\end{equation}
Given their embedding in an end-to-end trainable neural network architecture, both the instruments model $r$ and the renderer $c$ are required to be differentiable, in order to allow the error signal to be backpropagated through them. Recently, following the introduction by Kato et al. \cite{kato2018neural} of a differentiable 3-D mesh renderer, specifically designed for neural networks, some works have successfully introduced renderers in their pipelines, as in \cite{wu2020unsupervised}. In this work, we address these challenges by directly learning the mapping between the kinematic vector $\boldsymbol{\hat{k}}$ and the output binary mask $\hat{m}$. Given a geometric forward kinematic model of the instruments and a renderer (both non-necessarily differentiable), one can generate a virtually infinite set of coupled samples $\{\boldsymbol{k_i},m_i\}$ and use them to train a neural network to learn the direct mapping $\rho = r \odot c: \boldsymbol{\hat{k}} \rightarrow \hat{m}$.
The network is implemented with a generator-like architecture, consisting of a first fully connected layer, whose output is reshaped and fed to a fully convolutional network, alternating normal convolution and deconvolution layers. The last layer of the network has a sigmoid activation function, and outputs an image $\hat{m}$, with same dimension as $x$ and a single channel.\\
The training loss is formulated as a binary cross-entropy loss:
\begin{equation}
    \mathcal{L}_j = -\frac{1}{N}\sum_{i=1}^{w\cdot h}\left[m_i^j log(\hat{m}_i^j) + (1-m_i^j) log(1-\hat{m}_i^j)\right],
\end{equation}
with $\hat{m}_i$ being the predicted value for pixel $i$, and $w,h$ being the mask dimensions.

\subsection{Regressor and Decoder}
\label{section:encdec}
The regressor $\psi$ processes the input image $x$ to predict the corresponding instruments kinematic configuration $\boldsymbol{\hat{k}}$. Given the general independence between instruments (if more than one are simultaneously present), $\psi$ is implemented using a shared backbone and $m$ separate heads, one per robotic instrument. The backbone is based on ResNet50 \cite{boroumand2018deep}, using only the first and second convolutional layers (21 residual blocks). Each head consists of four convolutional layers, followed by global average pooling and a three-layers fully connected network, having $n/m$ units in output. In order to take advantage of the rough information provided by the measured kinematics $k_s$, without using it as a strong supervision signal, we introduce a ``soft mean squared error'' auxiliary loss function $\Loss_a$, defined as:
\begin{equation}
    \Loss_a = \frac{1}{n}\sum_{i=1}^{n}max((k_i-\hat{k}_i)^2-t^2, 0),
\label{eq:lossa}
\end{equation}
with $t$ being an hyperparameter, which we refer to as ``tolerance'', and that can be interpreted as a maximum accepted offset of $\boldsymbol{\hat{k}}$ with respect to $\boldsymbol{k_s}$. The idea behind this loss is to ease the optimization process by providing the network with a range in which the solution is likely to be found, avoiding any hard-coded constraint.

The decoder $\phi$ processes the predicted projection mask $\hat{m}$ and the \textit{backgroundized} image $x_b$ to reconstruct the image $\hat{x}$. The network is implemented as a \textit{UU-net}, an extension of the well-established \textit{U-net} architecture \cite{ronneberger2015u}, having two separate contracting paths for $\hat{m}$ and $x_b$, and an expanding path where corresponding features from the two contracting branches are concatenated. In order to learn the auto-encoder formulation defined in eq. \ref{eq:autoencod}, enforcing the reconstructed image $\hat{x}$ to match the input image $x$, we implemented  a perceptual loss $\Loss_r$, defined as:
\begin{equation}
    \Loss_r = \lVert \Gamma(x)-\Gamma(\hat{x})\rVert^2,
\label{eq:lossr}
\end{equation}
where $\Gamma$ is a feature extractor implemented as a VGG-16 network \cite{simonyan2014very} pre-trained on the publicly available Imagenet dataset \cite{deng2009imagenet}. Compared to pixel-wise losses, the perceptual loss has shown more robust results in many image reconstruction tasks \cite{bruna2015super}\cite{dosovitskiy2016generating}.

\begin{figure}[!tpb]
    \centering
      \includegraphics[width=3in]{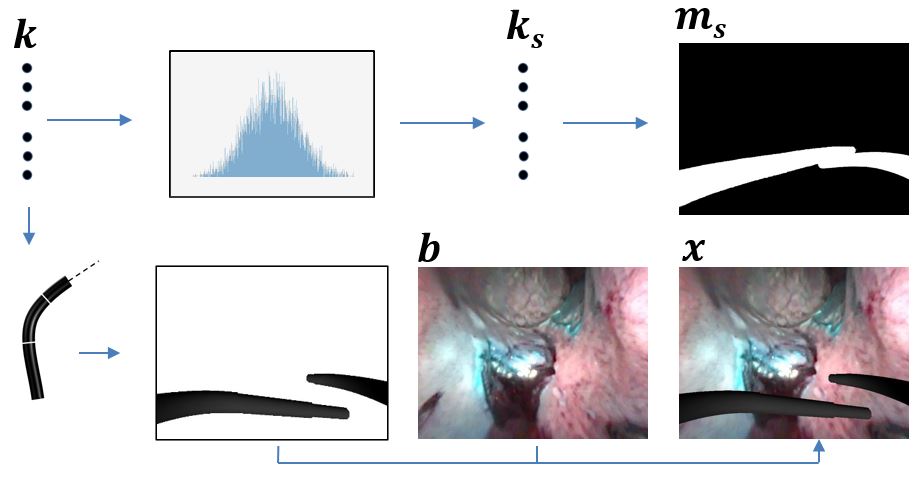}
      \caption{Semi-synthetic dataset building: a kinematic configuration $\boldsymbol{k}$ and a background $b$ are randomly sampled. $\boldsymbol{k}$ is then fed to a VTK model of the robotic instruments and rendered on the image space. The projection is then blended with $b$ to obtain the image $x$. Parallelly, a random noise is added to $k$ to simulate the measured kinematics $\boldsymbol{k_s}$ (visualized as the corresponding binary projection $m_s$ on the image plane, in order to qualitatively show the difference with the GT kinematic configuration $\boldsymbol{k}$).}
      \label{fig:semi-synth}
\end{figure}

\section{EXPERIMENTAL VALIDATION}
\label{exp_valid}
The experimental validation was performed using the STRAS robot \cite{de2013introducing}, a teleoperated prototype for flexible robotic endoscopic surgery \cite{legner2017endoluminal}. The robot is built as a standard endoscope having two operating instrument channels, through which robotic arms can be positioned. The robot arms used in STRAS are flexible cable-actuated instruments. Each instrument has 3 joint angles/positions, resulting in a total dimension of $\boldsymbol{\hat{k}}$ equal to $6$: rotation around the instrument main axis, translation along the same axis, and a cable-actuated bending, defined by the delta between the lengths of the two cables used for actuation. The instruments main axes are parallel to the exit parts of the instrument channels, which deviate from the endoscope main axis, forming a small angle ($\sim\ang{10}$), resulting almost perpendicular to the camera plane. During working configuration, instruments are bent, in order to be inside the field-of-view of the camera; therefore, when the rotational joint is activated, the visible part of the instruments moves on a plane almost parallel to the camera plane, avoiding ambiguities between pose and projected shape \cite{allan20183}. 
\newline Using the constant curvature assumption, the joint angles can be used to compute the forward kinematic model of the robotic instruments, following equations detailed in \cite{webster2010design}. The robot instruments are teleoperated by the user sending commands through a master console. The positions of the motors are recorded into the array of joint values $\boldsymbol{k_s}$. Camera images are acquired through an acquisition board in a synchronized fashion, resulting in RGB images with $570\times760$ resolution. The geometric forward kinematic model of the instruments and the renderer were implemented using the VTK library. The two VTK models were used exclusively in the datasets generation process, as detailed later in this paragraph, and not included in the training architecture, given their non-differentiability.\\
\indent Three different datasets were used for validating the proposed approach, two from real acquisitions and a semi-synthetic one. The first real dataset (\textit{phantom}) was recorded on the benchtop, using a plastic phantom model of the human digestive system. During the acquisition both the endoscope and the plastic model were moved, in order to avoid having a static background appearance. The second real dataset  (\textit{in-vivo}) was recorded during a 4-day set of \textit{in-vivo} experiments on porcine models \footnote{The study protocol for this experiment was approved by the Institutional Ethical Committee on Animal Experimentation (ICOMETH No.38.2011.01.018). Animals were managed in accordance with French laws for animal use and care as well as with the European Community Council directive no. 2010/63/EU}. This dataset presents several challenges compared to standard retinal, laparoscopic and benchtop datasets, including low foreground/background contrast, highly cluttered and changing background, frequent occlusions and bleedings, and strong tool-tissue interactions. The \textit{semi-synthetic} dataset was created by blending background-only images, automatically extracted from the \textit{in-vivo} dataset by parsing the associated kinematic information (although imprecise), with rendered robot instruments, obtained using the VTK model of tools and renderer, and random kinematic configurations $\boldsymbol{k}$. In order to simulate a realistic imprecise kinematic information $\boldsymbol{k_s}$ for this dataset, we added to the nominal value $\boldsymbol{k}$ a normally distributed noise, whose range was empirically chosen to match the real-dataset noise. The full process followed to build the semi-synthetic dataset is shown in fig. \ref{fig:semi-synth}.\\
\indent For the \textit{semi-synthetic} dataset, the ground truth (GT) joint configuration is known a-priori, due to the way the dataset was built.  For the \textit{phantom} and \textit{invivo} datasets, the GT joint configuration is unknown, since every measurement attempt is affected by uncertainties, resulting on the imprecise estimation $\boldsymbol{k_s}$. Instead, the evaluation on these two datasets was performed indirectly: the 3-D shape of the instruments is reconstructed according to the predicted kinematics $\boldsymbol{\hat{k}}$, and projected on the camera plane: the projected mask $\hat{m}$ can be then compared to the instruments ground truth location on the images, obtained via manual segmentation, providing an indirect evaluation of the estimated $\boldsymbol{\hat{k}}$.\\ \indent In order to train and evaluate the \textit{physical module}, a dataset was built generating random couples $\{\boldsymbol{k_i},m_i\}$, using the VTK model of instruments and renderer. For the training of the inpainting network, two separate datasets were built. The first one, \textit{real backgrounds}, was built automatically extracting background-only images from the \textit{in-vivo} dataset, according to the associated kinematic information (although imprecise),  and used to train the \textit{backgroundizer} for the \textit{semi-synthetic} and \textit{in-vivo} experiments; the second one, \textit{phantom backgrounds} was built by extracting a single background-only image from the \textit{phantom dataset}, strongly augmented through operations like rotation, cropping, lighting etc., and used for the \textit{phantom} experiment.
Table \ref{table:data} summarizes the number of images and GT images for the  datasets.

\begin{table}
\caption{Number of images in each training and testing dataset (* GT 2-D segmentation mask only, obtained via manual segmentation)}
\centering
\begin{tabular}{c||c|c}
    \textbf{Dataset}  &  \textbf{\# Images Training} & \textbf{\# GT testing images}\\
    \hline
    \hline
    \textit{physical module} &   \phantom{2}100k &  10k\\
    \hline
    \textit{real backgrounds} &   \phantom{2}6000 &  / \\
    \hline
    \textit{phantom backgrounds} &   \phantom{2}1 (+augmentation) &  / \\
    \hline
    \textit{semi-synthetic}   &  20400 & 2400 \\
    \hline
    \textit{phantom} &   \phantom{1}6800 &  \phantom{1}800*\\
    \hline
    \textit{in-vivo} & 28800 (4 days) &  \phantom{1}400*/day \\
    \hline
\end{tabular}
\label{table:data}
\end{table}

\begin{table*}\centering
\caption{Semi-synthetic dataset results. Comparison with raw kinematics $k_s$ and fully supervised methods $BSupK_s$, $BSupSoftK_s$. For the joint mean absolute errors (translation: \textit{t.}, rotation: \textit{r.}, bending: \textit{b.}), lower is better. For the reprojection metrics (intersection over union: \textit{IoU}), higher is better.}
\begin{tabular}{@{}lccccccccccc@{}}
\toprule
& \multicolumn{4}{c}{$left$} & \phantom{a}& \multicolumn{4}{c}{$right$}\\
\cmidrule{2-5} \cmidrule{7-10}
& $t.[mm]$ & $r.[deg]$ & $b.[mm]$ & \phantom{1}$IoU$\phantom{12} && $t.[mm]$ & $r.[deg]$ & $b.[mm]$ & \phantom{1}$IoU$\phantom{12}\\
\hline \\
$\boldsymbol{k_s}$ & 6.40 & 16.40 & 0.87 & 0.244 && 3.21 & 13.07 & 0.77 & 0.422\\
$BSupK_s$ & 4.00 & 10.30 & 0.73 & 0.646 && 1.85 & 7.05 & 0.49 & 0.646\\
$BSupSoftK_s$ & 4.27 & 11.00 & 0.62 & 0.355 && 2.13 & 6.97 & 0.50 & 0.620\\
$ours$ & \textbf{1.75} & \textbf{6.02} & \textbf{0.47} & \textbf{0.738} && \textbf{1.17} & \textbf{3.61} & \textbf{0.30} & \textbf{0.854}\\
\bottomrule
\end{tabular}
\label{tab:semi-synth}
\end{table*}

\begin{table}\centering

\caption{Evaluation of the IoU on the real datasets (Phantom \& in-vivo). Comparison with raw kinematics $k_s$ and fully supervised methods $BSupK_s$, $BSupSoftK_s$. For the in-vivo the average of the results obtained for each day is reported.}
\begin{tabular}{@{}lccccc@{}}
\toprule
& \multicolumn{2}{c}{$left$} & \phantom{a}& \multicolumn{2}{c}{$right$}\\
\cmidrule{2-3} \cmidrule{5-6}
& \phantom{4}$phant.$\phantom{6} & \phantom{1}$in\textnormal{-}vivo$\phantom{4} && \phantom{4}$phant.$\phantom{4} & $in\textnormal{-}vivo$\\
\hline \\
$\boldsymbol{k_s}$ & 0.283 & 0.421 && 0.327 & 0.447\\
$BSupK_s$ & 0.280 & 0.482 && 0.327 & 0.461\\
$BSupSoftK_s$ & 0.313 & 0.458 && 0.330 & 0.436\\
$\textit{ours}$ & \textbf{0.640} & \textbf{0.554} && \textbf{0.725} & \textbf{0.554}\\
\bottomrule
\end{tabular}

\label{tab:real_results}
\end{table}

\subsection{Training Details}
\label{train_det}
As a preliminary step, the inpainting network, belonging to the \textit{backgroundizer module} $\beta$, and the neural network implementation of the \textit{physical module} $\rho$ were trained. The \textit{physical module} was trained on the randomly generated couples $\{\boldsymbol{k_i},m_i\}$, for $100$ epochs, using a batch size of $64$ and a learning rate of $0.0005$. For the inpainting network the grid resolution was set to $6\times8$, as experimentally found providing the best trade-off between area covered and inpainting quality. Two inpainting networks were trained (on \textit{real backgrounds} and \textit{phantom backgrounds}), with a batch size of $12$ and a learning rate of $0.001$, until visually satisfying results were reached.\\
The two modules were then frozen during the end-to-end training of the regressor $\psi$ and the decoder $\phi$. The regressor and the decoder were trained alternatively (1 iteration each) using the Adam optimizer with learning rates equal to $0.0005$ and $0.00001$, respectively, a batch size of $36$, for 100 epochs: the decoder is trained on loss $\Loss_r$; the regressor is trained on a weighted sum of losses $\Loss_r$ and $\Loss_a$, defined as:

\begin{equation}
    \Loss = \alpha_a \Loss_a + \alpha_r \Loss_r,
\label{eq:loss}
\end{equation}

with $\alpha_a$, $\alpha_r$ being weight parameters set to $10$ and $0.001$, respectively, in order to balance the magnitude of the two losses.
The tolerance parameter $t$ was tuned through an hyperparameter search, adjusting a nominal value empirically determined in the context of other works using the STRAS robotic system \cite{da2019self}\cite{de2013introducing}.

\begin{figure*}[t]
  \includegraphics[width=\textwidth]{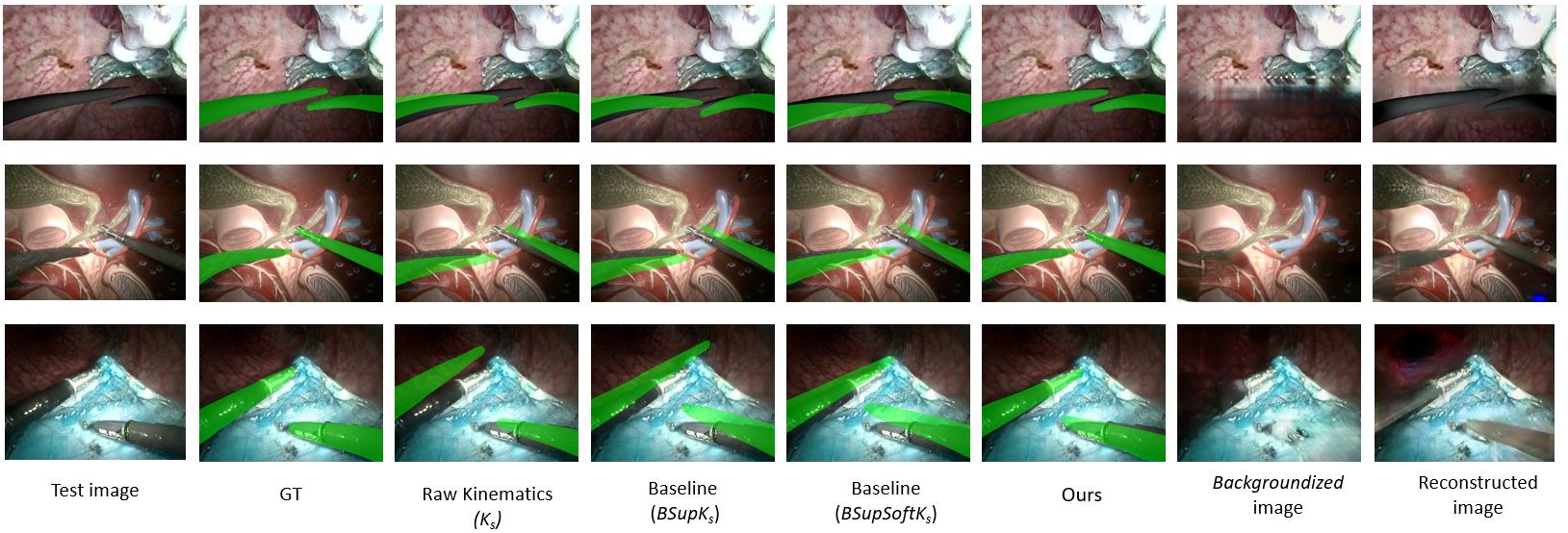}
  \caption{Examples of qualitative results on the three testing datasets. Top row: \textit{semi-synthetic}; middle row: \textit{phantom}; bottom row: \textit{in-vivo}. The last two columns show corresponding reconstructed image and \textit{backgroundized} image. Note that at inference time none of the two is needed/obtained, since the image-based regressor $\psi$ is a completely independent module.}
  \label{fig:qualit}
\end{figure*}

\begin{figure}[tpb]
    \centering
      \includegraphics[width=2.81in]{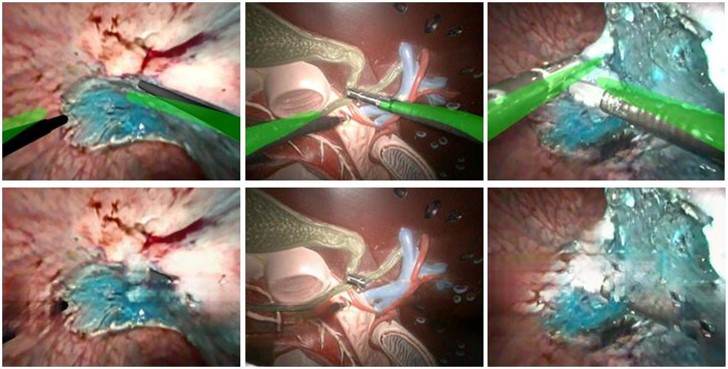}
      \caption{Examples of \textit{backgroundizer} module results. Top row shows one image for each dataset (\textit{semi-synthetic},\textit{phantom}, \textit{in-vivo}), with the corresponding measured kinematics $k_s$ projection (imprecise). Bottom row shows the corresponding \textit{backgroundized} versions, obtained using the rough localization provided by $k_s$.}
      \label{fig:back_res}
\end{figure}

\section{RESULTS}
In this section we present the validation results obtained by the proposed approach on the three datasets. In order to assess the actual contribution of the method, we compared its performances to two baseline models. They both consist of solely the regressor model $\psi$ trained by minimizing only the loss function $\Loss_a$ defined in eq. \ref{eq:lossa}. The first baseline model $\textit{BSupK}_s$ was trained by setting the tolerance parameter $t$ to $0$, thus resulting in a strongly-supervised training, having $\boldsymbol{k_s}$, which is unreliable/imperfect, as target; the second baseline model $\textit{BSupSoftK}_s$ was trained by setting the parameter $\boldsymbol{t}$ to the same value as the one used for our main model, thus resulting in a softer supervision. Together with the baselines, also the raw measured kinematics $\boldsymbol{k_s}$ provided by the robotic system was evaluated, according to the same modalities.\\
For the evaluation on the \textit{in-vivo} dataset, leave-one-out cross-validation was performed, by training the models on 3-days and testing on the remaining. The average of the results across the 4 days was then computed and reported.

\subsection{Backgroundizer and physical modules preliminary evaluation}

The \textit{physical module}, preliminary trained, was evaluated resulting in a mean Intersection over Union (IoU) of $94.3\%$ on the testing dataset. The inpainting networks (\textit{backgroundizer} module), preliminary trained, were qualitatively evaluated. Qualitative results for the three datasets are shown in fig. \ref{fig:back_res}.

\subsection{Regressor evaluation}
For the \textit{semi-synthetic} dataset, the GT kinematic configuration $\boldsymbol{k}$ of the instruments is known, and the mean absolute error for each joint could be computed. Reprojection of the estimated instruments shape on the image plane was also evaluated through the standard segmentation metric Intersection over Union (IoU). Results are reported in table \ref{tab:semi-synth}. For the real \textit{phantom} and \textit{in-vivo} datasets the reprojection error with respect to the manually annotated GT was evaluated, via IoU. Results are reported in table \ref{tab:real_results}. Qualitative results for the three datasets can be observed in fig. \ref{fig:qualit}.\\
\indent Results on the three datasets, despite not reaching a state-of-the-art accuracy, show a consistent improvement provided by the proposed approach. This is shown directly by the improved kinematics estimation in the \textit{semi-synthetic} dataset, and indirectly by the higher reprojection accuracies on the \textit{semi-synthetic} and real datasets, with respect to both the baselines and the raw kinematics. 
In the \textit{semi-synthetic} and \textit{in-vivo} datasets the baselines perform better than the raw kinematics $\boldsymbol{k_s}$, while in the \textit{phantom} dataset the performances are comparable. This may be due to the smaller size of the \textit{phantom} dataset, which prevents the networks from learning to average out the random noise of $\boldsymbol{k_s}$ with respect to the GT value $\boldsymbol{k}$. Our model, on the contrary, is not affected, proving its capability to learn also from reduced amount of noisy data. The evaluation of $\textit{BSupK}_s$ provides an ablation study of our method, showing that the improvements reached are not solely caused by the soft loss. In the \textit{in-vivo} dataset, a lower improvement is reached. This can be explained by two main reasons: 1) the dataset is more challenging due to the factors mentioned in section \ref{exp_valid} (e.g. low instruments-background contrast), which can affect our model more than the baselines, since the training is almost completely image-based; 2) phenomena as tool-channel interaction and slackening \cite{camarillo2008mechanics}, more evident under strong tool-tissue interaction, can deform the flexible instruments in configurations that our model, based on constant curvature assumption, cannot reach. An additional interesting aspect of the proposed method is that the fully trained architecture can be divided in autonomous, fully-functional sub-modules:
\begin{itemize}
    \item the regressor $\psi$, the core element of the method,
    \item a segmentation module, combination of $\psi$ and $\rho$,
    \item an image synthesizer $\phi$, able to generate a real-looking sample from  a background image and a GT segmentation mask of the tools (fig. \ref{fig:qualit}, last column).
\end{itemize}
Finally, the average processing time for each image is approximately 30 ms ($\sim$ 33 fps) on a single Tesla V100 GPU.

\section{CONCLUSION AND FUTURE WORK}

In this work, we propose a method for 3-D pose estimation of flexible surgical instruments. The proposed approach is self-supervised and trains a model to regress the kinematic configuration of the instruments directly from surgical camera images. In order to train this model with no manual annotations, the problem is formulated in an auto-encoder framework, by exploiting a tight \textit{kinematic bottleneck} to force a regressor to learn the actual kinematics. Our approach also takes advantage, at training time only, of imprecise kinematic information provided by the robotic system along with the camera images. While the kinematic data is imprecise due to tool-tissue interactions and is usually difficult to access in real-time applications inside the operating room, it helps drive the learning process to a strong regressor that estimates even better kinematics. One strong advantage is that no manual annotation is required during training and no additional optimization, is performed at inference time. Furthermore, in addition to the regressed kinematics, the model directly provides a segmentation mask of the tools. 

Experimental evaluation was carried out using a particular robotic endoscope having two flexible robotic arms, each one characterized by three joint values, on three datasets: \textit{semi-synthetic}, \textit{phantom} and \textit{in-vivo}. Results show a consistent improvement with respect to the baselines trained directly on the raw instruments kinematics.

\indent While the results are promising, the proposed approach presents a few limitations that could be addressed. First, the kinematic model of the instruments should be improved, in order to take into account phenomena as tool-channel interactions and slackening, particularly evident in non-controlled environments, as the \textit{in-vivo} dataset. Secondly, the uncertainty associated with the robot kinematics could be explicitly modelled, instead of relying on the naïve grid-approach, with possible benefits on the \textit{backgroundizer} and the whole training process. Finally, temporal dependencies between consecutive images could be modelled.




\section*{ACKNOWLEDGMENT}
Thanks to Bernard Dallemagne, Florent Nageotte, and Philippe Zanne for the help in generating the datasets.\\
This version is the preprint of an article to be published in IEEE Robotics and Automation Letters, \href{https://doi.org/10.1109/LRA.2021.3062308}{DOI 10.1109/LRA.2021.3062308}. Copyright 2021, IEEE.

\bibliography{bibliography} 
\bibliographystyle{ieeetran}

\end{document}